\documentclass{article}

 \usepackage[preprint]{neurips_2026}

\usepackage[utf8]{inputenc} 
\usepackage[T1]{fontenc}    
\usepackage{hyperref}       
\usepackage{url}            
\usepackage{booktabs}       
\usepackage{amsfonts}       
\usepackage{nicefrac}       
\usepackage{microtype}      
\usepackage{xcolor}         
\usepackage{graphicx}
\usepackage{tikz}
\usetikzlibrary{shapes.geometric, arrows.meta, positioning, fit, backgrounds}
\usepackage{amsmath}
\usepackage{cleveref}
\usepackage{fancyvrb}
\usepackage{tcolorbox}
\usepackage{float}
\usepackage{amsthm}
\newtheorem{definition}{Definition}
\usepackage{algorithm}
\usepackage{algorithmic}
\usepackage{multirow} 
\usepackage{caption}
\usepackage{enumitem}
\usepackage{tabularx}
\usepackage{booktabs}

\title{CausalFlow: Causal Attribution and Counterfactual Repair for LLM Agent Failures}

\author{%
  Akash Bonagiri\thanks{Correspondence to \texttt{sbonagiri@ucdavis.edu}.} \quad
  Devang Borkar \quad
  Gerard Janno Anderias \\
  \textbf{Setareh Rafatirad} \quad
  \textbf{Houman Homayoun} \\
  Department of Computer Science \\
  University of California, Davis \\
  Davis, CA 95616 \\
}

\begin{document}

\maketitle

\begin{abstract}
Large language model (LLM) agents frequently fail on multi-step tasks involving reasoning, tool use, and environment interaction. While such failures are typically logged or retried heuristically, they contain structured signals about where execution broke down. We introduce \textbf{CausalFlow}, an interventional framework that converts failed agent traces into minimal counterfactual repairs and reusable supervision. CausalFlow models execution traces as sequential chains of dependent steps and computes \emph{Causal Responsibility Scores} (CRS) via step-level counterfactual intervention to identify failure-inducing steps. For these steps, we generate minimally edited repairs that flip the final outcome to success, producing validated contrastive pairs of the form (wrong step, corrected step). CausalFlow supports two complementary uses: targeted test-time repair that recovers from failures with minimal behavioral drift, and training-time supervision suitable for offline preference optimization or reward modeling. Across four benchmarks spanning mathematical reasoning, code generation, question answering, and medical browsing, CausalFlow converts failed executions into validated minimal repairs with high minimality and causal-consensus scores, and demonstrates that causal attribution is necessary for reliable improvement across diverse agent tasks, outperforming heuristic refinement in complex retrieval settings while producing more localized repairs throughout. These results demonstrate that interventional analysis over structured execution traces provides a principled and scalable mechanism for transforming agent failures into reliability gains and learning-ready supervision.
\end{abstract}

\section{Introduction}

Large language model (LLM) agents execute complex multi-step procedures involving reasoning, tool invocation, and environment interaction \citep{liu2024agentbench, zhou2023webarena, guo2024stabletoolbench, wang2024voyager, yao2022react}. When such executions fail, supervision is typically restricted to outcome-level feedback (e.g., final answer correctness), leaving ambiguous which intermediate decision caused the failure \citep{lightman2024verify}. This ambiguity limits both reliability and learning: without principled credit assignment over execution traces, failures cannot be systematically repaired or converted into structured supervision.

Prior approaches address agent failures through heuristic retries, critique-driven rewriting, or full solution regeneration \citep{shinn2023reflexion, madaan2023selfrefine, yao2023tree, gou2024critic, chen2024selfdebugging, yang2024sweagent}. While effective in some cases, these strategies do not explicitly test causal responsibility within structured traces. As a result, they neither isolate the failure-inducing step nor guarantee that repairs correspond to minimal counterfactual corrections.

We introduce \textbf{CausalFlow}, a framework for interventional analysis and counterfactual repair of multi-step agent executions. CausalFlow models an execution trace as a sequential chain of dependent steps, where each step's output propagates forward to downstream computations. Given a failed trace, it performs step-level counterfactual interventions: replacing a candidate step and re-executing downstream computation to test whether the final outcome changes. This defines a \emph{Causal Responsibility Score (CRS)}, identifying steps whose intervention flips failure to success.

\begin{figure*}[t]
\centering
\resizebox{0.90\textwidth}{!}{%
\begin{tikzpicture}[
    node distance=0.8cm and 1.0cm,
    mainbox/.style={rectangle, draw, rounded corners, minimum height=1.4cm, minimum width=2.4cm, align=center, font=\small\bfseries, line width=1pt},
    subbox/.style={rectangle, draw, rounded corners=2pt, minimum height=0.6cm, minimum width=2.4cm, align=center, font=\scriptsize},
    inputbox/.style={rectangle, draw, rounded corners, minimum height=1.0cm, minimum width=1.8cm, align=center, font=\small, dashed},
    outputbox/.style={cylinder, shape border rotate=90, draw, minimum height=1.4cm, minimum width=1.4cm, shape aspect=0.25, align=center, font=\small, line width=1.5pt},
    arrow/.style={-{Stealth[length=3mm]}, thick},
    note/.style={font=\scriptsize, align=center},
    label/.style={font=\scriptsize\itshape, text=gray},
]
    \node[inputbox, fill=gray!10] (input) {Failed Trace\\$\tau = (s_1, \ldots, s_n)$};

    \node[mainbox, fill=orange!15, right=of input] (crs) {Causal Step\\Identification};
    \node[subbox, fill=orange!10, below=0.2cm of crs] (crs1) {Intervene on each step $s_i$};
    \node[subbox, fill=orange!10, below=0.08cm of crs1, text width=2.6cm] (crs2) {Compute CRS$(s_i) = \mathbb{I}[\mathcal{V}(\tau') = 1]$};

    \node[mainbox, fill=green!15, right=of crs] (repair) {Counterfactual\\Repair};
    \node[subbox, fill=green!10, below=0.2cm of repair] (rep1) {For each $s_i$ with CRS$=1$};
    \node[subbox, fill=green!10, below=0.08cm of rep1, text width=2.8cm] (rep2) {Generate repairs $s'_i$, rank by minimality};

    \node[mainbox, fill=blue!15, right=of repair] (validate) {Multi-Agent\\Validation};
    \node[subbox, fill=blue!10, below=0.2cm of validate] (val1) {Re-execute $\tau[i \leftarrow s'_i]$};
    \node[subbox, fill=blue!10, below=0.08cm of val1] (val2) {Keep if $\mathcal{V}(\tau') = 1$};

    \node[outputbox, fill=red!10, right=of validate] (output) {Validated\\Repairs};

    \draw[arrow] (input) -- (crs);
    \draw[arrow] (crs) -- (repair);
    \draw[arrow] (repair) -- (validate);
    \draw[arrow] (validate) -- (output);

    \node[note, below=0.25cm of output, text width=2.0cm] (outnote) {$(s_i, s'_i, \tau)$\\contrastive pairs};

\end{tikzpicture}
}
\caption{Overview of \textsc{CausalFlow} pipeline for causal attribution and counterfactual repair. Given failed execution trace, we identify causally responsible steps via interventional scoring (CRS), generate minimal counterfactual repairs \& validate repairs through re-execution or outcome prediction.}
\label{fig:pipeline}
\end{figure*}
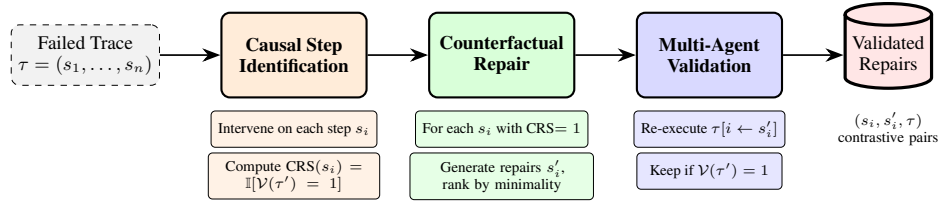

For causally responsible steps, CausalFlow generates minimally edited counterfactual repairs and validates them via deterministic re-execution or outcome prediction. Each validated repair yields a structured contrastive pair (wrong step, corrected step), transforming execution failures into reusable supervision (\Cref{fig:pipeline}). This enables deploy-time repair that recovers from failures without parameter updates. We evaluate CausalFlow across four benchmarks spanning mathematical reasoning (GSM8K), code generation (MBPP), question answering (SealQA Hard), and medical browsing (MedBrowseComp), totaling over 3,000 problems. CausalFlow converts 42.7\% of failed executions into validated minimal repairs and demonstrates measurable improvements in downstream task success through targeted counterfactual repair.

Our contributions are:
\begin{enumerate}[leftmargin=1.4em]
  \item \textbf{CausalFlow}: an interventional framework for step-level
  causal attribution in structured agent traces. Given a failed execution,
  CausalFlow replaces candidate steps and re-executes downstream computation
  to identify decisions whose intervention flips the final outcome.

  \item \textbf{Counterfactual repair}: a minimal repair mechanism that
  converts causally responsible failures into validated supervision pairs
  $(s_i, s_i^\star)$. These pairs support deploy-time repair and provide
  learning-ready contrastive examples for offline preference optimization
  or reward modeling.

  \item \textbf{Empirical validation}: evaluation across four domains,
  including mathematical reasoning, code generation, question answering,
  and medical browsing. Across more than 3{,}000 problems, CausalFlow
  converts 42.7\% of failed executions into validated minimal repairs and
  improves downstream task success through targeted intervention.

  \item \textbf{Failure-to-supervision formulation}: a general formulation
  for transforming logged execution failures into reusable step-level
  supervision, enabling offline learning from naturally occurring agent
  failures without requiring manual annotation of every intermediate step.
\end{enumerate}

\section{Related Work}

\textbf{Iterative refinement and self-repair.} A large body of recent work improves multi-step LLM agents through iterative refinement, self-reflection, and tool-assisted repair mechanisms. Reflexion introduces verbal reinforcement learning where agents critique prior outputs and store reflective feedback for future attempts \citep{shinn2023reflexion}. Self-Refine proposes iterative generation with model-produced feedback to improve responses without additional supervision \citep{madaan2023selfrefine}. Self-Reflection generates structured error diagnoses to guide re-answering
\citep{10852426}. ReAct interleaves reasoning traces with tool calls to enable interactive problem solving \citep{yao2022react}, while Tree-of-Thoughts expands reasoning through structured search over intermediate steps \citep{yao2023tree}, CRITIC incorporates external tools to validate and revise model outputs \citep{gou2024critic}, and recent self-debugging approaches enable models to iteratively repair generated code using execution feedback \citep{chen2024selfdebugging}. Agentic systems such as SWE-agent further scale iterative debugging to large software repositories \citep{yang2024sweagent}. Although these methods improve reliability through critique loops, search, or execution feedback, they generally rely on heuristic refinement or full regeneration rather than explicitly identifying and validating the causally responsible step via counterfactual intervention. 

\textbf{Step-level supervision and preference learning.} Alignment methods such as RLHF train reward models from human preference comparisons and optimize policies using reinforcement learning \citep{ouyang2022training, kaufmann2023rlhfsurvey}, and recent preference optimization frameworks such as Direct Preference Optimization (DPO) \citep{rafailov2023dpo}, Ranking-based Reward Learning (RRHF) \citep{yuan2023rrhf}, Odds-Ratio Preference Optimization (ORPO) \citep{hong2024orpo}, and offline RL approaches such as Implicit Language Q-Learning (ILQL) \citep{snell2023ilql} aim to improve credit assignment and stability without full RL pipelines. Nonetheless, preference signals are typically defined at the trajectory or response level, leaving ambiguity about which intermediate step caused failure.

\textbf{Failure diagnosis and causal intervention.} MAST \citep{cemri2025multiagentllmsystemsfail}
taxonomizes 14 failure modes but neither intervenes nor repairs. Who\&When
\citep{zhang2025agentcausestaskfailures} benchmarks failure attribution and
finds state-of-the-art models achieve only 14.2\% step-level accuracy from
logs alone. DoVer \citep{ma2026doverinterventiondrivenautodebugging} is the
most closely related concurrent work, but targets a structurally distinct setting: multi-agent systems with explicit orchestrator/sub-agent topology. DoVer requires framework-level checkpoint and replay infrastructure, segments traces into planning-execution trials using re-plan steps as cut points, and first generates log-based attribution hypotheses before validating them through targeted intervention. CausalFlow differs along three axes: it targets single-agent sequential traces without requiring framework modification or checkpointing infrastructure; its Causal Responsibility Score is a purely interventional quantity computed by directly replacing candidate steps and propagating effects forward through affected descendants, bypassing the intermediate log-based hypothesis generation stage; and it explicitly ranks repairs by minimality and produces validated contrastive pairs $(s_i, s_i^\star)$ for offline preference optimization and reward modeling, a use case DoVer does not address . Causal tracing and model editing
in transformers \citep{meng2022rome, meng2023memit}, LLM causal reasoning studies
\citep{chi2024causal, kiciman2024causalllm}, and causal abstraction frameworks
\citep{geiger2025causalabstraction} apply interventional ideas to internal model
behavior rather than explicit execution traces.

\section{Problem Setup}

We consider multi-step LLM agents that solve tasks through structured execution traces containing reasoning steps, tool invocations, and environment observations.

\paragraph{Execution traces.}
For a task instance $x$, an agent produces
\[
\tau = (s_1, s_2, \dots, s_T),
\]
where each step $s_t$ contains an action, such as a reasoning token sequence or tool call, and, when applicable, an environment observation. Let $y(\tau)$ denote the induced task output, and let
\[
\mathcal{V}(y(\tau), x) \in \{0,1\}
\]
be a task-specific verifier indicating success or failure.

\paragraph{Dependency structure.}
Execution traces exhibit explicit step dependencies \citep{zhang2025graphtracer}. A step may consume outputs from earlier steps, such as intermediate variables, tool responses, or retrieved information. We model the trace as a sequential chain where each step $s_t$ depends on all preceding steps:
\[
    s_1, \ldots, s_{t-1}.
\]
Thus, intervening on $s_i$ requires re-executing all subsequent steps:
\[
    s_{i+1}, \ldots, s_T.
\]
In our implementation, dependencies are logged from the agent runtime: tool responses depend on their tool calls, reasoning steps depend on the most recent reasoning step and referenced observations, and environment observations depend on the preceding environment action.

\paragraph{Counterfactual intervention.}
Given a failed trace $\tau$ with $\mathcal{V}(y(\tau), x)=0$, we define a step-level intervention by replacing $s_t$ with an alternative $\tilde{s}_t$ and re-executing subsequent steps. The resulting trace is
\[
\tau[t \leftarrow \tilde{s}_t].
\]
An intervention is successful if
\[
\mathcal{V}(y(\tau[t \leftarrow \tilde{s}_t]), x) = 1.
\]

Our goal is to identify steps whose intervention flips failure to success and generate minimal counterfactual repairs that remain successful under re-execution.
\section{The CausalFlow Framework}

We present the four components of \textsc{CausalFlow}: trace modeling, causal attribution via interventional scoring, counterfactual repair, and multi-agent validation.

\subsection{Causal Attribution via Interventions}

Given a failed trace $\tau = (s_1, \ldots, s_n)$ with $\mathcal{V}(y(\tau), x)=0$, our goal is to identify which steps are \emph{causally responsible} for the failure. Inspired by interventionist accounts of actual causation \cite{halpern2005causes}, we treat a step as responsible if replacing it and propagating its downstream effects changes the final outcome.

\paragraph{Sequential intervention.}
To intervene on step $s_i$, we replace it with an alternative $s_i'$ and re-execute all subsequent steps $s_{i+1}, \ldots, s_T$, while keeping preceding steps unchanged. This localizes the intervention to the candidate failure-inducing step and its downstream consequences, avoiding full trace regeneration.

\begin{definition}[Causal Responsibility Score]
\label{def:crs}
For a failed trace $\tau$, we generate $K$ intervention proposals $\{s_i'^{(k)}\}_{k=1}^K$ for step $s_i$. The \emph{Causal Responsibility Score} is:
\begin{equation}
\label{eq:crs}
\mathrm{CRS}(s_i)
=
\max_{k \in \{1,\dots,K\}}
\mathbb{I}\!\left[
\mathcal{V}\!\big(y(\tau[i \leftarrow s_i'^{(k)}]), x\big)=1
\right],
\end{equation}
where $\tau[i \leftarrow s_i'^{(k)}]$ denotes sequential re-execution after intervening on $s_i$ and recomputing only affected descendants.
\end{definition}

Thus, $\mathrm{CRS}(s_i)=1$ if at least one replacement of $s_i$ flips the verifier outcome from failure to success.

\paragraph{Computing interventions and re-execution.}
For each step $s_i$, we prompt an LLM with the step content, dependency context, and failure feedback ~\citep{bonagiri2025towards} to generate $K$ minimally edited correction proposals. Reasoning-step interventions correct logical errors, while tool-call interventions adjust arguments or tool selection. Intervened traces are evaluated by deterministic re-execution when an executor exists, such as a Python interpreter, and by predictive re-execution otherwise. \Cref{alg:crs} summarizes CRS computation.

\subsection{Counterfactual Repair}

For steps with $\mathrm{CRS}(s_i)=1$, we generate validated \emph{counterfactual repairs}. To preserve interpretability and isolate the root cause, we rank successful interventions by minimality, measured through position-wise token matching with a length penalty:
\begin{equation}
    \label{eq:minimality}
    \text{Minimality}(s_i, s_i') =
    \frac{m}{L}\left(1 - \tfrac{1}{2}\cdot\frac{\left|\;|x|-|y|\;\right|}{L}\right),
\end{equation}
where $x=\text{tokens}(s_i)$ and $y=\text{tokens}(s_i')$ are token sequences, $L=\max(|x|,|y|)$, and $m=\sum_{k=1}^{\min(|x|,|y|)} \mathbb{I}[x_k = y_k]$ counts position-wise token matches. Higher scores indicate smaller edits.

Among successful interventions, we select the repair maximizing minimality:
\begin{equation}
\label{eq:repair}
\begin{aligned}
s_i^\star 
&= \arg\max_{s_i'} \ \text{Minimality}(s_i, s_i') \\
\text{s.t.} \quad 
&\mathcal{V}\!\big(y(\tau[i \leftarrow s_i']), x\big)=1 .
\end{aligned}
\end{equation}

Each validated repair yields a contrastive supervision pair $(s_i, s_i^\star)$.

\subsection{Multi-Agent Validation}
\label{sec:multi_agent_validation}

Because LLM-generated attributions may be noisy, we use a three-agent validation system \citep{liang2024mad,bonagiri2026stableval}: Agent A proposes causal steps, Agent B critiques the proposed attributions, and Agent C meta-critiques both judgments. Each critic returns an agreement label $\{\texttt{AGREE}, \texttt{PARTIAL}, \texttt{DISAGREE}\}$ and confidence score $c_j \in [0,1]$. We map agreement labels to $a_j \in \{1,0.5,0\}$ for \texttt{AGREE}, \texttt{PARTIAL}, and \texttt{DISAGREE}, respectively.

We define the consensus score as:
\begin{equation}
\label{eq:consensus}
\mathrm{Consensus}(s_i)
=
\frac{1}{3}
\Big(
\mathrm{CRS}(s_i)
+
\sum_{j \in \{B,C\}} c_j a_j
\Big).
\end{equation}

The factor $1/3$ ensures $\mathrm{Consensus}(s_i)\in[0,1]$. Unless otherwise specified, we set $\tau_c=0.5$ and retain steps with $\mathrm{Consensus}(s_i)\geq\tau_c$ as confirmed causal steps.

\section{Experiments}

We evaluate \textsc{CausalFlow} across four domains to assess whether it can identify causally responsible steps, generate validated counterfactual repairs, and improve task performance.

\subsection{Benchmarks}

We evaluate on four benchmarks spanning mathematical reasoning, program synthesis, web-based question answering, and medical browsing. GSM8K \cite{cobbe2021gsm8k} consists of grade-school math word problems requiring multi-step arithmetic reasoning. MBPP \cite{austin2021program} evaluates Python program synthesis with executable correctness checks. SealQA Hard \cite{pham2025sealqa} involves complex question answering requiring iterative web search and information synthesis. MedBrowseComp \cite{chen2025medbrowsecomp} evaluates medical question answering in a browsing environment where the agent must navigate external resources. Together, these datasets comprise over 3,000 task instances and vary in tool usage, reasoning depth, and environmental interaction complexity.

\subsection{Agent Configuration}
\label{sec:agent_config}

For each benchmark, we use a model suited to its tool requirements and representative of realistic deployment configurations. GSM8K uses Gemini 2.0 Flash Lite with a calculator tool; MBPP uses GPT-5 Chat with Docker-based test execution; and SealQA Hard and MedBrowseComp use Gemini 3 Flash Preview with web search via Serper API. All agents produce structured traces with explicitly logged step types and dependencies, a necessary precondition for CRS computation since plain chain-of-thought provides no identifiable intervention points. This structure introduces a modest accuracy cost relative to unstructured baselines, which the repair mechanism is designed to recover.

We compare \textsc{CausalFlow} against Direct single-pass chain-of-thought, Self-Refine \citep{madaan2023selfrefine}, and Self-Reflection \cite{10852426}. All baselines use the same model checkpoints and temperature settings as the corresponding \textsc{CausalFlow} agent. On GSM8K and MBPP, baselines refine their own generated answers; on SealQA Hard and MedBrowseComp, they share a single BrowseCompAgent web-search run and refine using only the collected context, isolating the refinement mechanism. 

\subsection{Intervention Protocol}

For each failed execution, \textsc{CausalFlow} computes CRS using $K=3$ intervention proposals per candidate step. Baselines do not perform step-level interventions and are evaluated only on their final outputs. Intervened traces are evaluated by deterministic re-execution when an executor is available, as in MBPP, and by predictive outcome modeling otherwise, as in GSM8K, SealQA Hard, and MedBrowseComp.

\subsection{Evaluation Metrics}

We report four metrics. \textbf{Repair Rate} measures the fraction of failed traces converted to correct executions after refinement or repair; Direct has zero repair rate by definition. \textbf{Post-Repair Accuracy} measures overall task accuracy after applying each method. \textbf{Minimality Score} measures repair localization using the position-wise token similarity formula from \Cref{eq:minimality}; higher scores indicate smaller edits. \textbf{CRS Precision} measures how often \textsc{CausalFlow}-flagged steps admit a validated outcome-flipping intervention.

\section{Results}

We now present empirical results evaluating repair effectiveness, overall performance impact, attribution precision, repair characteristics, and cross-domain behavior patterns across benchmarks.

\begin{table}[H]
\centering
\footnotesize
\setlength{\tabcolsep}{4pt}
\caption{%
  Repair performance across benchmarks and methods.
  \textit{Repair Rate} computed over failed traces only.
  \textit{Min.} = position-wise token similarity between original
  and repaired steps (higher $=$ smaller edits).
  Direct repair rate is 0\,\% by definition.%
}
\label{tab:repair_baseline}
\begin{tabular}{@{}llrrrcc@{}}
\toprule
\textbf{Benchmark} & \textbf{Method}
  & \textbf{Total} & \textbf{Pass} & \textbf{Fail}
  & \textbf{Repairs (\%)} & \textbf{Min.} \\
\midrule
\multirow{4}{*}{GSM8K}
  & Direct          & 1319 & 1162 & 157  & 0\,(0.0)    & 1.00 \\
  & Self-Refine     & 1319 & 1148 & 171  & 5\,(2.9)    & 0.99 \\
  & Self-Reflection & 1319 & 1155 & 164  & 35\,(21.3)  & 0.88 \\
  & \textbf{CausalFlow} & 1319 & 989 & 330 & \textbf{173\,(52.4)} & 0.87 \\
\midrule
\multirow{4}{*}{MBPP}
  & Direct          & 974 & 501 & 473 & 0\,(0.0)    & 1.00 \\
  & Self-Refine     & 974 & 496 & 478 & 384\,(80.3) & 0.68 \\
  & Self-Reflection & 974 & 501 & 473 & 267\,(56.4) & 0.65 \\
  & \textbf{CausalFlow} & 947 & 523 & 488 & \textbf{201\,(41.2)} & \textbf{0.82} \\
\midrule
\multirow{4}{*}{SealQA Hard}
  & Direct          & 254 &  93 & 161 & 0\,(0.0)   & 1.00 \\
  & Self-Refine     & 254 &  92 & 162 & 27\,(16.7) & 0.17 \\
  & Self-Reflection & 254 &  92 & 162 & 28\,(17.3) & 0.01 \\
  & \textbf{CausalFlow} & 254 & 108 & 146 & \textbf{32\,(21.9)} & \textbf{0.79} \\
\midrule
\multirow{4}{*}{MedBrowse}
  & Direct          & 484 & 170 & 314 & 0\,(0.0)    & 1.00 \\
  & Self-Refine     & 484 & 169 & 315 & 12\,(3.8)   & 0.82 \\
  & Self-Reflection & 484 & 169 & 315 & 21\,(6.7)   & 0.65 \\
  & \textbf{CausalFlow} & 484 & 149 & 335 & \textbf{149\,(44.5)} & \textbf{0.84} \\
\bottomrule
\end{tabular}
\end{table}

\subsection{Main Repair Performance}
\label{sec:main_repair_performance}

Table~\ref{tab:repair_baseline} summarizes repair performance across all methods and domains. CausalFlow converts between 21.9\% and 52.4\% of failed traces into validated successful executions, producing repairs for 555 out of 1299 failed executions (42.7\%) in aggregate. Direct achieves zero repairs by definition across all benchmarks. Self-Refine and Self-Reflection show low repair rates on GSM8K (2.9\% and 21.3\%), MedBrowseComp (3.8\% and 6.7\%), and SealQA Hard (16.7\% and 17.3\%), with CausalFlow exceeding both on all three benchmarks.

MBPP is the exception, where Self-Refine achieves a repair rate of 80.3\% and Self-Reflection 56.4\%, both exceeding CausalFlow's 41.2\%. However, minimality scores on MBPP tell a different story: Self-Refine scores 0.68 and Self-Reflection 0.65, compared to CausalFlow's 0.82, indicating that the higher repair rates come at the cost of substantially larger edits. On SealQA Hard this contrast is most pronounced, where Self-Reflection's minimality score of 0.01 indicates near-complete answer regeneration versus CausalFlow's 0.79.

GSM8K achieves the highest CausalFlow repair rate (52.4\%), reflecting that arithmetic reasoning failures are typically confined to a single miscalculated or mispropagated step. SealQA Hard shows the lowest (21.9\%), where many failures stem from retrieval gaps that local intervention cannot address. MedBrowseComp (44.5\%) and MBPP (41.2\%) fall in between, with minimality scores of 0.84 and 0.82 respectively indicating that successful repairs remain tightly localized to the failure-inducing step across both domains.

\subsection{Impact on Overall Accuracy}

Table~\ref{tab:accuracy_baseline} reports baseline and post-refinement accuracy across all methods. CausalFlow yields consistent accuracy improvements across all four benchmarks, and is the only method to do so. Self-Refine and Self-Reflection produce negative deltas on MedBrowseComp (-1.4pp and -3.5pp respectively), indicating that global critique-based refinement actively degrades performance in long browsing traces. CausalFlow achieves the largest absolute gain on MedBrowseComp (+30.8pp, from 30.8\% to 61.6\%) and SealQA Hard (+12.6pp, from 42.5\% to 55.1\%), the two retrieval-heavy benchmarks where baselines fail to improve.

On GSM8K, CausalFlow post-repair accuracy (88.1\%) matches Direct and falls 
just below Self-Reflection (90.2\%). This result should be interpreted in 
light of the architectural tradeoff structured trace logging imposes: the 
13.1pp accuracy gap between CausalFlow's initial accuracy (75.0\%) and the 
Direct baseline (88.1\%) reflects the cost of requiring the agent to produce 
typed, dependency-annotated steps rather than free-form chain-of-thought. 
Crucially, the repair mechanism fully recovers this gap, demonstrating that 
the structured logging overhead is not a permanent accuracy penalty but a 
recoverable cost that enables principled causal attribution. CausalFlow's 
post-repair accuracy matches Direct exactly, at a lower per-failure cost and 
with the added benefit of interpretable step-level attribution that Direct 
cannot provide.

\subsection{Minimality and Localization of Repairs}

Across all benchmarks, CausalFlow achieves average minimality scores between 0.79 and 0.87 (Table ~\ref{tab:repair_baseline}), indicating that successful repairs typically involve small token-level modifications rather than wholesale rewriting. Qualitative inspection confirms that most repairs modify only a single reasoning statement, tool argument, or conditional clause, and in the majority of repaired traces only one or two steps receive CRS=1.

Baseline minimality scores provide useful contrast  (Table ~\ref{tab:repair_baseline}). Scores range from 0.99 (Self-Refine on GSM8K) down to 0.01 (Self-Reflection on SealQA Hard), with the near-zero score indicating full answer regeneration rather than targeted correction. Self-Refine on MBPP scores 0.68, reflecting broad function rewrites that happen to pass tests rather than isolated logic fixes. CausalFlow's scores of 0.82 and 0.79 on MBPP and SealQA Hard respectively sit substantially above both baselines on those benchmarks, confirming that causal attribution produces more localized repairs even in domains where heuristic methods resort to large rewrites.

This localization property is most pronounced in GSM8K and MBPP, where arithmetic and logic errors occur at identifiable intermediate steps. In browsing tasks, repairs involve minimal modifications to query phrasing or inference transitions. Across all domains, CausalFlow isolates root-cause decisions while preserving unaffected reasoning structure, producing contrastive pairs that more precisely identify the causal decision than pairs derived from wholesale regeneration, making them better suited for downstream preference optimization and reward modeling.

\subsection{Error Pattern and Skill Decomposition}

Clustering causally responsible steps reveals interpretable skill deficits across domains.

In GSM8K, failures cluster into arithmetic operations, multi-digit multiplication, unit conversion, percentage calculations, constraint tracking, and equation setup errors. MBPP failures concentrate around loop construction, string manipulation, recursion handling, boundary conditions, type conversion, and input validation. SealQA Hard primarily involves query formulation errors, incomplete multi-source synthesis, and improper answer extraction. MedBrowseComp failures frequently involve medical terminology interpretation, symptom-condition mapping, navigation decisions, and reasoning over treatment guidelines. This decomposition suggests that causal attribution not only identifies failure-inducing steps but also exposes systematic weaknesses in domain-specific skills. Such signals could potentially guide curriculum-based training or targeted fine-tuning in future work.

\subsection{Ablation Studies}
\label{sec:ablations}

We conduct ablation experiments to evaluate key components; full results are
in Tables~\ref{tab:stoch-subset}--\ref{tab:nogold}.

\textbf{Intervention sample count.} Repair rates improve up to $K{=}3$, after
which gains diminish relative to cost, indicating limited stochastic exploration
suffices to identify most repairable failures.

\textbf{Minimality ranking.} Removing minimality ranking produces larger edits
without meaningfully changing repair rate, confirming minimality primarily
enhances localization rather than raw repair success.

\textbf{Minimality metric sensitivity.} Lexical and edit-distance selectors
agree on 88\% of steps with multiple successful proposals
(Tables~\ref{tab:min-pool} and~\ref{tab:min-agree}). Semantic cosine agreement
drops from 90\% on GSM8K to 52\% on MedBrowseComp, where it loses
discriminative power~\citep{chakraborti2024nlp4gov,badam2022aletheia}; edit distance is the more robust alternative for
browsing-heavy domains.

\textbf{Stochasticity.} Three independent runs on 12 GSM8K failed traces
(Tables~\ref{tab:stoch-subset} and~\ref{tab:stoch-runs}) yield repair-rate
std.\ dev.\ of 6.81pp and post-repair accuracy std.\ dev.\ of 1.63pp,
indicating stable outcomes under repeated sampling.

\textbf{LLM judge accuracy.} Manual audit of 30 repairs per browsing benchmark
(Table~\ref{tab:judge}) found precision of 90.9\% on SealQA Hard and 86.2\%
on MedBrowseComp, adjusting repair rates to 19.9\% and 38.4\% respectively.

\textbf{Gold reference in repair prompts.} Removing gold costs 21pp on SealQA Hard and 6pp on MedBrowseComp, while producing slight gains on MBPP (+6.4pp) 
and negligible change on GSM8K (+0.55pp) (Table~\ref{tab:nogold}). This pattern reflects 
a structural difference between task types: retrieval-heavy benchmarks lack 
deterministic verifiers that would otherwise constrain repair generation, 
making the gold reference a stronger signal when the search space of valid 
repairs is large and underspecified. On structured reasoning tasks with 
executable verifiers, the repair search space is sufficiently constrained 
that gold provides no additional benefit. The gold reference is provided 
as reference only and the repair prompt explicitly instructs the model not 
to use it directly, preserving the interventional validity of the generated 
repairs.
\vspace{-8pt}
\section{Discussion}
\label{sec:discussion}

Our results suggest that many LLM agent failures are localized rather than systemic. 
Across benchmarks, \textsc{CausalFlow} often identifies one or a small number of steps 
whose intervention flips the final outcome, supporting the view that failed traces 
contain usable step-level supervision rather than only outcome-level feedback. This is 
important because most agent debugging and refinement methods treat the failed trajectory 
as a whole, making it difficult to separate the actual failure-inducing decision from 
the surrounding correct reasoning.

The benefit of causal repair depends strongly on task structure. On closed-form tasks 
such as GSM8K and MBPP, heuristic refinement can recover many failures, sometimes with 
higher final accuracy, but often through broader rewrites. In these settings, 
\textsc{CausalFlow}'s main advantage is not always raw accuracy, but repair localization 
and the production of cleaner contrastive supervision pairs. In retrieval-heavy settings 
such as SealQA Hard and MedBrowseComp, global refinement is less reliable because it can 
discard correct intermediate reasoning or fail to distinguish retrieval gaps from reasoning 
errors. Targeted intervention avoids this behavior by modifying only the step whose 
replacement changes the outcome.

These findings suggest that \textsc{CausalFlow} is best viewed not as a universal 
replacement for refinement, but as a causal debugging layer for structured agent traces. 
Its value is highest when practitioners need to understand why an execution failed, 
repair the failure with minimal behavioral drift, or convert failed traces into reusable 
training signals. The localized repairs also provide interpretable skill signals, such as 
arithmetic propagation errors, boundary-condition mistakes, query formulation failures, 
and medical reasoning misinterpretations. Such signals may support targeted debugging, 
curriculum construction, or future preference and reward modeling from failed executions.

Additional discussion on localized failures, causal intervention as an operational tool, 
domain-dependent repairability, and interpretability signals is provided in 
Appendix~\ref{app:extended_discussion}.
\section{Conclusion}

We introduced \textsc{CausalFlow}, a framework for causal debugging of LLM agent execution 
traces. By intervening on individual steps and re-executing downstream computation, 
\textsc{CausalFlow} identifies failure-inducing decisions and generates minimal validated 
repairs. Across four benchmarks spanning mathematical reasoning, program synthesis, 
web-based question answering, and medical browsing, it converts a substantial fraction of 
failed traces into successful executions while producing localized contrastive pairs for 
future supervision.

The central finding is that many agent failures can be repaired without regenerating the 
entire trajectory. This makes causal intervention useful both as an inference-time 
reliability mechanism and as a way to extract structured supervision from naturally 
occurring failures. As LLM agents become longer-horizon, tool-mediated, and increasingly 
deployed in high-stakes workflows, identifying not only whether they fail but where and why 
they fail will be essential. \textsc{CausalFlow} provides one step toward more repairable, 
interpretable, and learning-ready agent systems.
\section*{Limitations}

Several limitations warrant consideration.

First, intervention quality depends on the LLM’s ability to generate meaningful corrective proposals. If proposed interventions fail to approximate valid alternatives, causally responsible steps may be under-identified.

Second, computing CRS requires re-execution of affected trace segments, introducing computational overhead. Although empirical results indicate that small values of $K$ are sufficient, scaling to very long or tool-intensive traces may require approximation strategies.

Third, retrieval-driven failures cannot be corrected through local reasoning modification when essential information is unavailable. Integrating retrieval-aware intervention mechanisms remains an open challenge.

Fourth, CausalFlow's structured trace-logging configuration reduces initial accuracy relative to plain chain-of-thought on GSM8K (75.0\% vs.\ 88.1\% for Direct). This gap is an inherent consequence of requiring agents to produce typed, dependency-annotated execution traces: the structured format constrains the model's output space in ways that free-form generation does not face. The 
repair mechanism fully recovers this gap on GSM8K, but the logging overhead 
represents a deployment cost that practitioners must weigh against the 
interpretability and repairability gains CausalFlow provides. Future work could explore lighter-weight trace formats or post-hoc dependency annotation that reduce this accuracy cost without sacrificing the intervention semantics CRS requires.

Finally, the quality of causal attribution depends on accurate dependency annotations. Incomplete or noisy sequential chains may weaken intervention semantics and attribution reliability.

\section*{Impact Statement}
\label{sec:Impact}

This paper presents \textsc{CausalFlow}, a framework for identifying and repairing causally responsible steps in LLM agent execution traces. The primary goal of this work is to improve the reliability and interpretability of multi-step AI systems. By enabling structured causal debugging, \textsc{CausalFlow} may contribute to safer deployment of LLM-based agents in domains such as education, programming assistance, information retrieval, and healthcare support. Identifying localized failure causes can facilitate human oversight, targeted correction, and more transparent system behavior. At the same time, systematic analysis of failure modes could potentially be misused to probe system weaknesses or optimize adversarial inputs. However, the techniques presented here do not introduce new attack capabilities beyond what is already possible through standard evaluation and stress-testing practices. Instead, the framework is intended to support robustness analysis and reliability improvement. Overall, we believe that advancing methods for structured failure diagnosis and repair promotes safer and more interpretable AI systems. We encourage future work to integrate such causal debugging tools into broader safety and alignment pipelines.

\bibliographystyle{plainnat}  
\bibliography{example_paper}

\appendix

\section*{Appendix}

\section{Implementation Details}
\label{app:Appendix_A}

This section provides additional details on trace logging, intervention prompting, and repair validation.

\subsection{Causal Responsibility Scoring Algorithm}

\begin{algorithm}[H]
\caption{Causal Responsibility Scoring}
\label{alg:crs}
\small
\begin{algorithmic}[1]
  \STATE \textbf{Input:} Failed trace $\tau = (s_1, \ldots, s_n)$, instance $x$, verifier $\mathcal{V}(\cdot, x)$
  \STATE \textbf{Output:} CRS scores $\{\text{CRS}(s_i)\}_{i=1}^{n-1}$
  \FOR{$i = 1$ \textbf{to} $n-1$}
    \STATE $\text{CRS}(s_i) \leftarrow 0$
    \STATE $\{s_i'^{(k)}\}_{k=1}^{K} \leftarrow \text{GenerateInterventions}(s_i,\, \tau_{<i},\, \text{feedback})$
    \FOR{$k = 1$ \textbf{to} $K$}
      \STATE $\tau' \leftarrow \text{SequentialReExec}(\tau,\, i,\, s_i'^{(k)})$
      \IF{$\mathcal{V}(y(\tau'), x) = 1$}
        \STATE $\text{CRS}(s_i) \leftarrow 1$
        \STATE \textbf{break}
      \ENDIF
    \ENDFOR
  \ENDFOR
  \STATE \textbf{return} $\{\text{CRS}(s_i)\}$
\end{algorithmic}
\end{algorithm}

\subsection{Trace Logging and Step Typing}

We implement a \textsc{TraceLogger} that records agent execution as a sequence of typed steps. Each step contains:

\begin{itemize}
    \item A discrete step type,
    \item Step-specific content (e.g., reasoning text, tool arguments, or observations),
    \item Explicit dependencies on prior steps.
\end{itemize}

The step types are defined as follows:

\begin{Verbatim}[fontsize=\footnotesize]
class StepType(Enum):
    REASONING = "reasoning"
    TOOL_CALL = "tool_call"
    TOOL_RESPONSE = "tool_response"
    LLM_RESPONSE = "llm_response"
    MEMORY_ACCESS = "memory_access"
    FINAL_ANSWER = "final_answer"
\end{Verbatim}

Each step maintains a list of dependency indices referencing earlier steps in the trace. These dependencies define the ordering of steps in the sequential trace.




\subsection{Intervention Prompting}
\label{app:intervention_prompting}

To generate candidate counterfactual interventions, we use a structured prompting template. For each candidate step, the LLM receives:

\begin{itemize}
    \item The trace context up to the intervened step,
    \item The failed step content,
    \item Feedback describing the failure outcome.
\end{itemize}

\begin{figure*}[t]
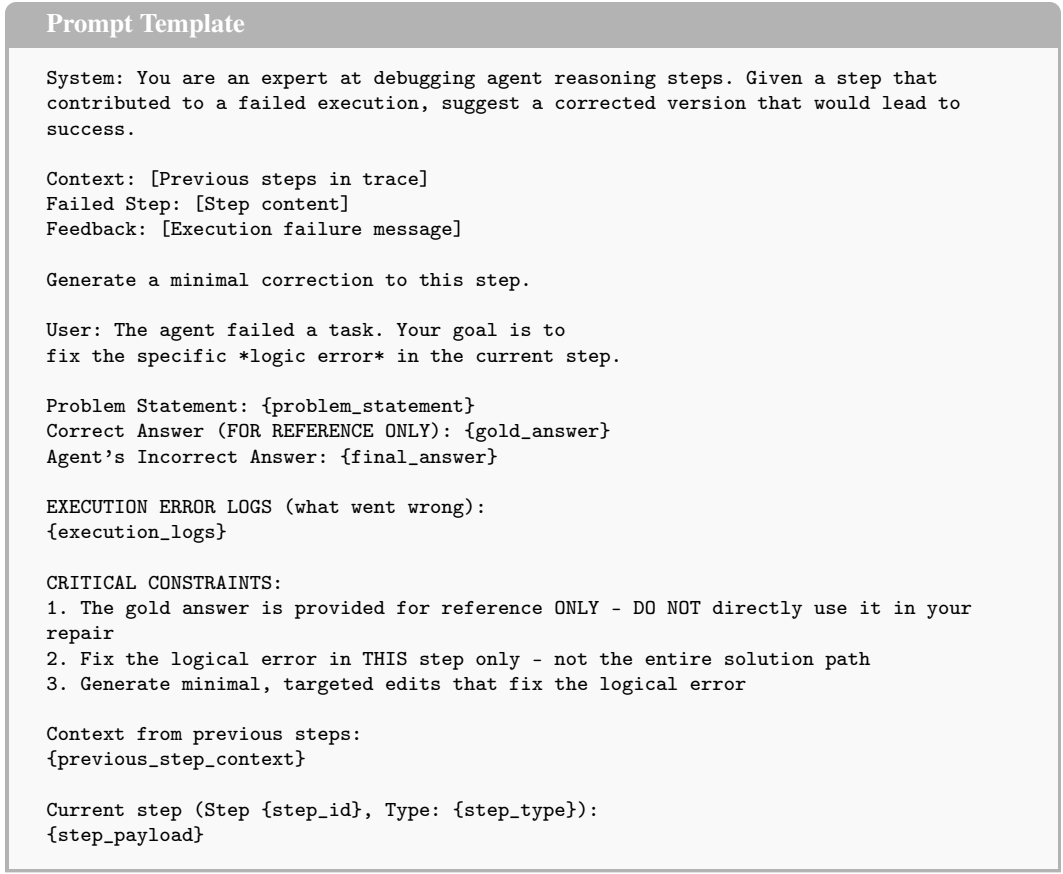

\centering
\begin{tcolorbox}[colback=gray!5, colframe=gray!60, title=\textbf{Prompt Template}, sharp corners=south]
    \begin{Verbatim}[fontsize=\footnotesize, commandchars=\\\{\}]
\textbf{System:} You are an expert at debugging agent reasoning steps. Given a step that 
contributed to a failed execution, suggest a corrected version that would lead to
success.

Context: [Previous steps in trace]
Failed Step: [Step content]
Feedback: [Execution failure message]

Generate a minimal correction to this step.

\textbf{User:} The agent failed a task. Your goal is to 
fix the specific *logic error* in the current step.

Problem Statement: \{problem_statement\}
Correct Answer (FOR REFERENCE ONLY): \{gold_answer\}
Agent's Incorrect Answer: \{final_answer\}

EXECUTION ERROR LOGS (what went wrong):
\{execution_logs\}

CRITICAL CONSTRAINTS:
1. The gold answer is provided for reference ONLY - DO NOT directly use it in your
repair
2. Fix the logical error in THIS step only - not the entire solution path
3. Generate minimal, targeted edits that fix the logical error

Context from previous steps:
\{previous_step_context\}

Current step (Step \{step_id\}, Type: \{step_type\}):
\{step_payload\}
    \end{Verbatim}
\end{tcolorbox}
\caption{Structured prompt template used to generate minimal counterfactual interventions for a candidate step. Note that variables in brackets are dynamically populated during the repair generation phase.}
\label{fig:intervention_prompt}
\end{figure*}

The prompt template is shown in \Cref{fig:intervention_prompt}.

Sampling is performed with temperature-based decoding to generate multiple intervention proposals per step.

\subsection{Additional Prompt Templates}

For completeness, we provide representative prompt templates used by the causal attribution component in \Cref{fig:causal_prompt_templates}. Braced fields denote instance-specific values.

\begin{figure*}[t]
\centering
\begin{tcolorbox}[colback=gray!5, colframe=gray!60, title=\textbf{Causal Attribution Prompt}, sharp corners=south]
    \begin{Verbatim}[fontsize=\footnotesize, commandchars=\\\{\}]
You are analyzing a failed agent execution. The agent produced an incorrect final
answer.

\textbf{Problem Statement:} \{problem_statement\}
\textbf{Gold Answer (correct answer):} \{gold_answer\}
\textbf{Environment feedback at the point of failure:} \{end_feedback\}

\textbf{Context from previous steps:}
\{previous_step_context\}

\textbf{Current step (Step \{step_id\}, Type: \{step_type\}):}
\{step_payload\}
    \end{Verbatim}
\end{tcolorbox}
\caption{Prompt used by CausalFlow for causal attribution. The model is tasked with identifying if the current step is the root cause of the failure.}
\label{fig:causal_prompt_templates}
\end{figure*}

\subsection{Repair Validation and Re-Execution}

Repairs are validated through either deterministic re-execution or predictive outcome estimation, depending on task availability of an executor.

For tasks such as MBPP and GSM8K, deterministic execution ensures reliable validation. For browsing-based tasks, outcome prediction relies on LLM-based evaluation of the modified trace.

\subsection{GSM8K Experimental Setup}
\label{app:GSM8K_setup}

\paragraph{Dataset.}
We use the GSM8K dataset \cite{cobbe2021gsm8k} loaded from HuggingFace (\texttt{gsm8k}, \texttt{main} config, \texttt{test} split), comprising 1,319 grade-school math word problems.

\paragraph{Model.}
\texttt{google/gemini-2.0-flash-lite-001} with temperature 0.7.

\paragraph{Agent Configuration.}
The GSM8K agent (\texttt{GSM8KAgent}) decomposes problems into structured steps: initial reasoning, LLM-generated structured solution with reasoning and calculation steps, calculator tool calls for each operation, and final answer extraction.

\paragraph{Tools.}
Calculator tool for evaluating mathematical expressions using \texttt{MathReexecutor}.

\paragraph{Prompts.}
The solve prompt requests step-by-step solutions with: (1) brief reasoning about approach, (2) each calculation step with description, operation type, and exact expression, (3) the final numerical answer.

\paragraph{Validation Method.}
LLM-based outcome prediction. Gold answers are extracted as numeric values using \texttt{MathReexecutor.extract\_number()}.

\subsection{MBPP Experimental Setup}

\paragraph{Dataset.}
We use the MBPP dataset \cite{austin2021program} loaded from HuggingFace (\texttt{Muennighoff/mbpp}), with train/test/validation/prompt splits merged. Total problems: 974.

\paragraph{Model.}
GPT-5 Chat with temperature 0.2 for code generation.

\paragraph{Agent Configuration.}
The MBPP agent reuses the HumanEval-style code generator (\texttt{HumanevalAgent}): reasoning step, LLM code generation tool call, generated Python code, Docker code execution tool call, test results, and final pass/fail answer.

\paragraph{Tools.}
\texttt{llm\_code\_generation} for generating Python code via LLM, and \texttt{docker\_code\_execution} for running code in isolated Docker containers with official test cases.

\paragraph{Prompts.}
Code generation prompt requests complete Python functions with exact function names, preserved signatures/docstrings, required imports, and no extra output.

\paragraph{Validation Method.}
Deterministic Docker-based execution using \texttt{HumanevalReexecutor}. Tests run in isolated containers; success requires all test assertions to pass.

\subsection{SealQA Hard Experimental Setup}
\label{app:sealQA_setup}

\paragraph{Dataset.}
We use SealQA Hard \cite{pham2025sealqa} loaded from HuggingFace (\texttt{vtllms/sealqa}, \texttt{seal\_hard} config, \texttt{test} split), comprising 254 complex QA problems requiring web search.

\paragraph{Model.}
\texttt{google/gemini-3-flash-preview} with temperature 0.0 for solver, 0.3 for agent steps.

\paragraph{Agent Configuration.}
The BrowseComp agent (\texttt{BrowseCompAgent}) uses a structured tool policy with maximum 15 steps. Actions include: \texttt{search}, \texttt{open\_url}, \texttt{extract}, and \texttt{answer}. State tracking maintains gathered facts, search history, and page summaries.

\paragraph{Tools.}
\texttt{web\_search} via Serper API with caching, and \texttt{web\_fetch} for fetching and parsing web pages.

\paragraph{Prompts.}
Agent system prompt instructs the model to search for relevant information, open promising URLs, extract key facts, and provide answers when sufficient information is gathered. LLM-based grader compares extracted final answers against gold answers.

\paragraph{Validation Method.}
LLM-based grading. Web results are cached for reproducibility.

\subsection{MedBrowseComp Experimental Setup}
\label{app:medbrowse_setup}

\paragraph{Dataset.}
We use MedBrowseComp \cite{chen2025medbrowsecomp} loaded from HuggingFace (\texttt{AIM-Harvard/MedBrowseComp\_CUA}), comprising 484 medical QA problems.

\paragraph{Model.}
\texttt{google/gemini-3-flash-preview} with temperature 0.3 for agent steps.

\paragraph{Agent Configuration.}
Same as SealQA (\texttt{BrowseCompAgent}) with maximum 10 steps (reduced for medical domain).

\paragraph{Tools.}
Same as SealQA: \texttt{web\_search}, \texttt{web\_fetch}.

\paragraph{Validation Method.}
LLM-based grading using the same grader template as SealQA.

\subsection{Runtime Analysis}
\label{app:runtime_analysis}
All LLM calls were made via the OpenRouter API\cite{openrouter}.
Runtime includes baseline trace generation, $K=3$
counterfactual interventions per candidate step,
sequential re-execution, and validation. Total wall-clock runtime and 
approximate API cost were: GSM8K (12 hours, \$31.56), MBPP (6 hours, 
\$9.72), SealQA Hard (1 hour, \$16.36), and MedBrowseComp (2 hours, 
\$27.05). Ablation experiments (Appendix~B.6) incurred an additional 
\$45.31, bringing the total API cost to approximately \$130.00. 
GSM8K used LLM-based outcome prediction for validation, while MBPP 
employed deterministic execution. Repair is applied only to failed 
traces; therefore, computational overhead scales with failure rate 
rather than dataset size. The average API cost per successful repair 
ranged between approximately \$0.05 and \$0.15 across datasets.

\section{Future Work}
Several promising directions follow from this work.

Validated counterfactual pairs $(s_i, s_i^\star)$ provide step-level supervision that could be used for targeted fine-tuning, reward modeling, or offline preference learning without requiring manual annotation. Structured causal monitoring could also be integrated into online agent execution, enabling self-repair before final output generation.

Extending causal debugging to multi-agent systems may reveal interaction-level failure modes and coordination errors. Additionally, learning to approximate CRS directly without full re-execution could reduce computational cost and enable scalable deployment.

More broadly, combining causal intervention with retrieval optimization or memory augmentation may address failure modes that are currently outside the scope of local reasoning repair.

Finally, reducing the accuracy overhead of structured trace logging 
remains an open challenge. Lighter-weight trace formats or 
post-hoc dependency annotation that preserves intervention semantics 
without constraining the model's output space would lower the deployment 
barrier for practitioners adopting CausalFlow in production settings.

\section{Extended Results}

\subsection{Per-Method Accuracy Before and After Refinement}

\begin{table}[H]
\centering
\footnotesize
\setlength{\tabcolsep}{4pt}
\caption{%
  Pre- and post-refinement accuracy per method.
  \textit{B} = Before (initial accuracy); \textit{A} = After
  (post-refinement). Direct: $B{=}A$ (no refinement).
  Negative $\Delta$ = performance degradation.%
}
\label{tab:accuracy_baseline}
\begin{tabular}{@{}l cc cc cc cc@{}}
\toprule
& \multicolumn{2}{c}{\textbf{GSM8K}}
& \multicolumn{2}{c}{\textbf{MBPP}}
& \multicolumn{2}{c}{\textbf{SealQA}}
& \multicolumn{2}{c}{\textbf{MedBrowse}} \\
\cmidrule(lr){2-3}\cmidrule(lr){4-5}
\cmidrule(lr){6-7}\cmidrule(lr){8-9}
\textbf{Method} & B & A & B & A & B & A & B & A \\
\midrule
Direct
  & .881 & .881 & .514 & .514 & .366 & .366 & .351 & .351 \\
Self-Refine
  & .870 & .870 & .509 & \textbf{.903} & .362 & .374 & .349 & .335 \\
Self-Reflection
  & .876 & .902 & .514 & .789 & .362 & .370 & .349 & .314 \\
\midrule
\textbf{CausalFlow}
  & .750 & \textbf{.881} & .552 & .764 & .425 & \textbf{.551} & .308 & \textbf{.616} \\
\midrule
\multicolumn{9}{@{}l}{\textit{Absolute gain $\Delta$ (After $-$ Before)}} \\[2pt]
Direct
  & \multicolumn{2}{c}{.000}
  & \multicolumn{2}{c}{.000}
  & \multicolumn{2}{c}{.000}
  & \multicolumn{2}{c}{.000} \\
Self-Refine
  & \multicolumn{2}{c}{.000}
  & \multicolumn{2}{c}{\textbf{+.394}}
  & \multicolumn{2}{c}{+.012}
  & \multicolumn{2}{c}{$-$.014} \\
Self-Reflection
  & \multicolumn{2}{c}{+.027}
  & \multicolumn{2}{c}{+.274}
  & \multicolumn{2}{c}{+.008}
  & \multicolumn{2}{c}{$-$.035} \\
\textbf{CausalFlow}
  & \multicolumn{2}{c}{\textbf{+.131}}
  & \multicolumn{2}{c}{+.212}
  & \multicolumn{2}{c}{\textbf{+.126}}
  & \multicolumn{2}{c}{\textbf{+.308}} \\
\bottomrule
\end{tabular}
\end{table}
\subsection{Visual Summary of CausalFlow results}

\begin{figure}[H]
  \centering
  \includegraphics[width=\columnwidth]{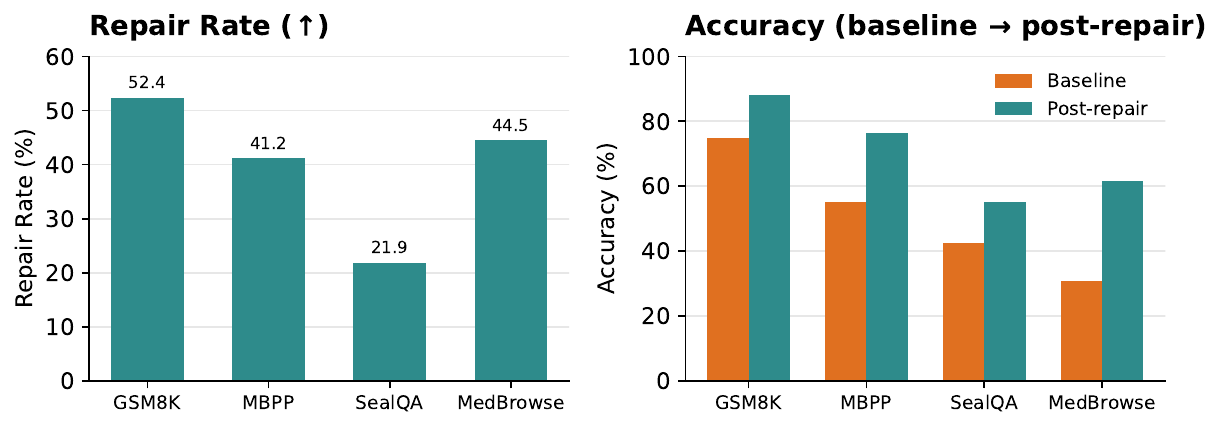}
  \caption{Visual summary of CausalFlow results across all four benchmarks.}
  \label{fig:results_combined}
\end{figure}

\subsection{Per-Model Breakdown}

Table~\ref{tab:per_model} provides a breakdown of repair performance by base model used for agent execution.

\begin{table}[H]
\centering
\footnotesize
\setlength{\tabcolsep}{4pt}
\caption{Repair performance broken down by base model.}
\label{tab:per_model}
\begin{tabular}{@{}llrrr@{}}
\toprule
\textbf{Benchmark} & \textbf{Model} & \textbf{Failed} & \textbf{Repaired} & \textbf{Repair Rate} \\
\midrule
GSM8K         & Gemini 2.0 Flash & 330 & 173 & 52.4\% \\
MBPP          & GPT-5 Chat       & 488 & 201 & 41.2\% \\
SealQA Hard   & Gemini 3 Flash   & 146 &  32 & 21.9\% \\
MedBrowse     & Gemini 3 Flash   & 335 & 149 & 44.5\% \\
\bottomrule
\end{tabular}
\end{table}

These results indicate that repair effectiveness is influenced more by task structure than by the specific base model.

\subsection{Dataset and Trace Statistics}

Table~\ref{tab:dataset_stats} provides detailed statistics for each benchmark, including baseline accuracy and repair outcomes.

\begin{table}[H]
\centering
\footnotesize
\setlength{\tabcolsep}{4pt}
\caption{Dataset and trace statistics. \textbf{Total / Avg.} row shows
         sums for counts and macro-averages for rates. Accuracy = Passed/Total.}
\label{tab:dataset_stats}
\begin{tabular}{@{}lrrrrr@{}}
\toprule
\textbf{Benchmark} & \textbf{Total} & \textbf{Passed} & \textbf{Failed} & \textbf{Fixed} & \textbf{Accuracy} \\
\midrule
GSM8K         & 1320 &  989 &  330 & 173 & 75.0\% \\
MBPP          &  947 &  523 &  488 & 201 & 55.2\% \\
SealQA Hard   &  254 &  108 &  146 &  32 & 42.5\% \\
MedBrowse     &  484 &  149 &  335 & 149 & 30.8\% \\
\midrule
\textbf{Total / Avg.} & \textbf{3004} & \textbf{1769} & \textbf{1299} & \textbf{555} & \textbf{50.8\%} \\
\bottomrule
\end{tabular}
\end{table}

\subsection{Skill Group Details}

Causal clustering reveals domain-specific skill categories. The following skill categories organize failure patterns for targeted debugging and potential curriculum learning. Category counts are derived from clustering causal steps using LLM-based skill labeling.

\paragraph{GSM8K Categories (16):}
Basic arithmetic, multi-digit multiplication, division with remainders, unit conversion, percentage calculation, ratio and proportion, multi-step planning, constraint satisfaction, word problem parsing, variable tracking, comparison operations, time calculations, money calculations, distance/rate/time reasoning, combinatorics basics, and equation setup.

\paragraph{MBPP Categories (12):}
Task decomposition and planning, algorithmic logic implementation, tool and API integration, code generation and syntactic integrity, requirement interpretation and mapping, data structure and type management, edge case and boundary handling, mathematical and geometric reasoning, regex and pattern synthesis, input parsing and verification, computational efficiency optimization.

\paragraph{SealQA Hard Categories (4):}
Query precision and formulation, source selection and extraction efficiency, iterative search and refinement, and temporal/contextual grounding.

\paragraph{MedBrowseComp Categories (6):}
Source relevance and authority assessment, iterative query refinement and narrowing, search query optimization and formulation, information extraction and parsing, redundancy and execution efficiency, and identifier/resource verification.

\paragraph{Using the Taxonomy.}
These categories enable: (1) identification of systematic skill deficits in agent behavior, (2) targeted debugging of specific failure modes, and (3) potential curriculum-based training strategies that progress from simpler to more complex skills within each domain.

\subsection{Causal Attribution Reliability Metrics}

In addition to repair success, we report two attribution-reliability metrics that evaluate whether \textsc{CausalFlow} identifies the correct failure-inducing step(s), and how often multi-agent validation agrees with the proposer.

\begin{table}[H]
\centering
\footnotesize
\setlength{\tabcolsep}{4pt}
\caption{Attribution-reliability metrics. CRS Precision measures how often
         CRS-flagged steps admit a validated outcome-flipping intervention.
         Consensus Rate measures the fraction of CRS-flagged steps retained
         after multi-agent validation. Multi-agent consensus was computed only
         for GSM8K; other benchmarks use deterministic re-executors that provide
         definitive success/failure outcomes directly.}
\label{tab:attrib_metrics}
\begin{tabular}{@{}lcc@{}}
\toprule
\textbf{Benchmark} & \textbf{CRS Precision} ($\uparrow$) & \textbf{Consensus Rate} ($\uparrow$) \\
\midrule
GSM8K       & 0.72 & 0.65 \\
MBPP        & 0.84 & ---  \\
SealQA Hard & 0.68 & ---  \\
MedBrowse   & 0.78 & ---  \\
\bottomrule
\end{tabular}
\end{table}

\paragraph{CRS Precision.}
For each failed trace $\tau$ with verifier outcome $\mathcal{V}(y(\tau),x)=0$, let
\[
\mathcal{S}_{\mathrm{pred}}(\tau) = \{ i \in \{1,\dots,n-1\} \;:\; \mathrm{CRS}(s_i)=1 \}
\]
be the set of steps flagged as causally responsible by CRS. Since ground-truth causal steps are not directly observed, we operationalize a conservative notion of correctness using repair validation: a predicted causal step is counted as a \emph{true positive} if there exists at least one intervention proposal for that step that flips the verifier outcome to success under sequential re-execution.
Formally, define
\[
\mathrm{TP}(\tau) = \big|\{ i \in \mathcal{S}_{\mathrm{pred}}(\tau) \;:\; \exists k,\ \mathcal{V}(y(\tau[i \leftarrow s_i'^{(k)}]),x)=1 \}\big|.
\]
Then CRS Precision is
\[
\mathrm{Prec}_{\mathrm{CRS}} = 
\frac{\sum_{\tau \in \mathcal{F}} \mathrm{TP}(\tau)}
{\sum_{\tau \in \mathcal{F}} |\mathcal{S}_{\mathrm{pred}}(\tau)|},
\]
where $\mathcal{F}$ is the set of failed traces.

\subsection{Ablation Studies}

\begin{table}[H]
\centering
\footnotesize
\setlength{\tabcolsep}{4pt}
\caption{GSM8K subset statistics for stochasticity ablation (seed=42).}
\label{tab:stoch-subset}
\begin{tabular}{@{}lrrrr@{}}
\toprule
\textbf{Dataset} & \textbf{Tot.} & \textbf{Pass} & \textbf{Fail} & \textbf{Acc.} \\
\midrule
GSM8K & 50 & 38 & 12 & 76.0\% \\
\bottomrule
\end{tabular}
\end{table}

\begin{table}[H]
\centering
\footnotesize
\setlength{\tabcolsep}{4pt}
\caption{Per-run repair results ($T$=0.7, $K$=3, multi-agent critique disabled).}
\label{tab:stoch-runs}
\begin{tabular}{@{}lrrrr@{}}
\toprule
\textbf{Run} & \textbf{Failed} & \textbf{Rep.} & \textbf{Rate} & \textbf{Post Acc.} \\
\midrule
Run 1 & 12 & 7 & 58.3\% & 90.0\% \\
Run 2 & 12 & 8 & 66.7\% & 92.0\% \\
Run 3 & 12 & 6 & 50.0\% & 88.0\% \\
\midrule
\textbf{Mean}       & & & 58.3\% & 90.0\% \\
\textbf{Std.\ Dev.} & & &  6.81\% &  1.63\% \\
\bottomrule
\end{tabular}
\end{table}

\begin{table}[H]
\centering
\footnotesize
\setlength{\tabcolsep}{4pt}
\caption{Minimality metric ablation: proposal pool statistics ($K$=5).
         Mean $\pm$ std characterizes metric behavior across the full proposal pool.}
\label{tab:min-pool}
\begin{tabular}{@{}lrrrrl@{}}
\toprule
\textbf{Benchmark} & \textbf{Traces} & \textbf{w/ repair} & \textbf{Steps} & \textbf{Props} & \textbf{lex / edit / sem} \\
\midrule
GSM8K       & 100 & 96  & 825  & 4125 & $0.70{\pm}0.33$ / $0.85{\pm}0.18$ / $0.97{\pm}0.04$ \\
MBPP        & 100 & 89  & 141  &  705 & $0.56{\pm}0.45$ / $0.68{\pm}0.34$ / $0.94{\pm}0.08$ \\
SealQA Hard &  26 & 22  &  65  &  325 & $0.29{\pm}0.25$ / $0.58{\pm}0.19$ / $0.92{\pm}0.06$ \\
MedBrowse   &  67 & 57  & 144  &  720 & $0.18{\pm}0.20$ / $0.50{\pm}0.20$ / $0.90{\pm}0.08$ \\
\midrule
\textbf{Total} & \textbf{293} & \textbf{234} & \textbf{1175} & \textbf{5875} & \\
\bottomrule
\end{tabular}
\end{table}

\begin{table}[H]
\centering
\footnotesize
\setlength{\tabcolsep}{4pt}
\caption{Minimality metric ablation: selector agreement, computed only on steps
         with $\geq$2 successful proposals (Multi-succ), where metric choice
         affects repair selection.}
\label{tab:min-agree}
\begin{tabular}{@{}lrrr@{}}
\toprule
\textbf{Benchmark} & \textbf{Multi-succ} & \textbf{Lex$=$Sem} & \textbf{Lex$=$Edit} \\
\midrule
GSM8K       & 615 & 90\% & 96\% \\
MBPP        &  82 & 73\% & 84\% \\
SealQA Hard &  48 & 54\% & 58\% \\
MedBrowse   & 106 & 52\% & 58\% \\
\midrule
\textbf{Total} & \textbf{851} & \textbf{82\%} & \textbf{88\%} \\
\bottomrule
\end{tabular}
\end{table}

\begin{table}[H]
\centering
\footnotesize
\setlength{\tabcolsep}{4pt}
\caption{Judge-accuracy audit for LLM-graded benchmarks. Unclear cases are
         excluded from precision computation. Adjusted rate = Reported rate
         $\times$ Precision. Wilson 95\% CIs reported for precision.}
\label{tab:judge}
\begin{tabular}{@{}lrrrrccc@{}}
\toprule
\textbf{Benchmark} & \textbf{Reviewed} & \textbf{Unclear} & \textbf{Scored}
  & \textbf{Precision} & \textbf{95\% CI} & \textbf{Rep.\ Rate} & \textbf{Adj.\ Rate} \\
\midrule
SealQA Hard & 30 & 8 & 22 & 90.9\% & [72.2\%, 97.5\%] & 21.9\% & 19.9\% \\
MedBrowse   & 30 & 1 & 29 & 86.2\% & [69.4\%, 94.5\%] & 44.5\% & 38.4\% \\
\bottomrule
\end{tabular}
\end{table}

\begin{table}[H]
\centering
\footnotesize
\setlength{\tabcolsep}{4pt}
\caption{No-gold repair-prompt ablation. Per-proposal success rates are strictly
         paired on traces where both arms produced proposals ($K$=5).
         $\Delta$ = No-gold $-$ With-gold (percentage points).
         $^*$Substituted from \texttt{gemini-2.0-flash-lite-001} due to
         upstream rate-limit failures.}
\label{tab:nogold}
\begin{tabular}{@{}llrrrr@{}}
\toprule
\textbf{Benchmark} & \textbf{Model} & \textbf{Paired Traces}
  & \textbf{With-Gold} & \textbf{No-Gold} & \textbf{$\Delta$ (pp)} \\
\midrule
GSM8K       & \texttt{llama-3.1-8b-instruct}$^*$  & 98 & 60.87\% & 61.42\% & $+0.55$ \\
MBPP        & \texttt{gpt-5-chat}                  & 68 & 54.26\% & 60.63\% & $+6.37$ \\
SealQA Hard & \texttt{gemini-3-flash-preview}       & 21 & 62.62\% & 41.38\% & $-21.24$ \\
MedBrowse   & \texttt{gemini-3-flash-preview}       & 61 & 64.68\% & 58.70\% & $-5.99$ \\
\bottomrule
\end{tabular}
\end{table}

\subsection{Code and Reproducibility}
\label{app:code_and_reproduce}

The complete implementation of \textsc{CausalFlow}, including all agent configurations, causal attribution algorithms, repair generation modules, and evaluation scripts, is publicly available for reproducibility and further research. The codebase includes detailed documentation, example traces, and instructions for running experiments across all four benchmarks evaluated in this work. Access to the anonymized repository is provided at: \url{https://anonymous.4open.science/r/CausalFlow-6B2C}.

\section{Extended Discussion}
\label{app:extended_discussion}
\subsection{Localized Nature of Agent Failures}

A central finding across benchmarks is that many agent failures are localized 
rather than systemic. In a large fraction of failed traces, only one to three 
steps receive CRS~$= 1$, indicating that a small number of decisions causally 
determine the incorrect outcome. This pattern holds across both structured 
reasoning domains, where arithmetic or logic mistakes occur at identifiable 
intermediate steps, and browsing-based environments, where failures frequently 
arise from localized reasoning transitions or query formulation decisions rather 
than complete reasoning collapse.

\subsection{Causal Intervention as an Operational Tool}
Unlike heuristic self-debugging approaches that rewrite reasoning based solely on global feedback, \textsc{CausalFlow} models traces as sequential chains and re-executes all subsequent steps from the intervention point forward, reducing behavioral drift and preserving unaffected reasoning components. The empirical gains observed across domains demonstrate that causal attribution is not merely explanatory but operational. Identifying and modifying specific failure-inducing steps leads to measurable improvements in overall task accuracy without modifying model parameters. This indicates that structured causal debugging can serve as a practical reliability mechanism at inference time.

\subsection{Domain-Dependent Repairability}
Repair effectiveness varies across domains, and the baseline comparisons reveal that this variation reflects a fundamental difference between task types rather than a property of CausalFlow alone. On closed-domain tasks with deterministic evaluators, all methods improve accuracy to some degree. Self-Refine and Self-Reflection achieve positive deltas on GSM8K and MBPP, and on MBPP Self-Refine reaches the highest final accuracy overall. In these settings, broad heuristic rewrites are sufficient to recover many failures because errors are discrete and the verifier is unambiguous. CausalFlow's advantage here is in repair precision rather than final accuracy. In retrieval-heavy environments the picture changes substantially. On SealQA Hard and MedBrowseComp, Self-Refine and Self-Reflection fail to improve reliably, with both producing negative deltas on MedBrowseComp. Global critique-based refinement cannot distinguish between reasoning errors and retrieval gaps, and risks discarding correct reasoning alongside incorrect conclusions. CausalFlow's localized intervention avoids this by targeting only the step whose replacement flips the verifier outcome, producing the largest accuracy gains in the paper on both benchmarks while baselines decline. This pattern suggests that causal attribution becomes increasingly necessary as task complexity and environmental interaction depth grow: in simple closed-form settings any refinement helps, but in long multi-step agentic settings only targeted causal repair improves performance consistently.

\subsection{Interpretability and Skill Signals}
Beyond performance improvement, \textsc{CausalFlow} provides interpretable insights into systematic agent weaknesses. Clustering causally responsible steps reveals domain-specific skill deficits, including arithmetic propagation errors in GSM8K, boundary-condition logic errors in MBPP, query formulation mistakes in SealQA Hard, and medical reasoning misinterpretations in MedBrowseComp. These patterns suggest that causal attribution exposes structured behavioral weaknesses rather than isolated random errors. Such signals could inform targeted training strategies, curriculum design, or structured reward shaping, enabling more efficient improvement than generic fine-tuning.

\section{Baseline Prompt Templates}
\label{sec:baseline_prompts}

For completeness and reproducibility, we provide the prompt templates used by the
Self-Refine and Self-Reflection baselines. All baselines use identical model
checkpoints and temperature settings to their corresponding \textsc{CausalFlow}
agent. On GSM8K and MBPP, prompts operate over the model's own generated
solution; on SealQA Hard and MedBrowseComp, all baselines share a single
\textsc{BrowseCompAgent} web-search run and condition refinement on the
collected context only.

\vspace{0.5em}
\noindent
\begin{minipage}{\textwidth}
\begin{tcolorbox}[colback=gray!5, colframe=gray!60,
    title=\textbf{Self-Refine Prompt Templates},
    sharp corners=south, width=\textwidth]
    \begin{Verbatim}[fontsize=\footnotesize, commandchars=\\\{\}]
\textbf{-- Stage 1: Generate --}
\textbf{System:} You are a careful math solver. Show all steps clearly.
\textbf{User:} Solve the following math problem step by step, showing every calculation.

Problem: \{problem\}

Solution:

\textbf{-- Stage 2: Feedback --}
\textbf{System:} You are a strict math reviewer. Find all errors precisely.
\textbf{User:} You are a rigorous math reviewer.

Problem: \{problem\}

Solution:
\{solution\}

Identify any errors in logic, arithmetic, or problem interpretation above.
Be specific: quote the wrong part, explain the mistake, and suggest the fix.
If the solution is completely correct, write '[STOP]' as the final line.

Feedback:

\textbf{-- Stage 3: Refine --}
\textbf{System:} You are a careful math solver. Fix all identified errors.
\textbf{User:} Revise the math solution below using the feedback provided.
Fix every identified error while keeping correct parts.

Problem: \{problem\}

Previous Solution:
\{solution\}

Feedback:
\{feedback\}

Revised Solution:
    \end{Verbatim}
\end{tcolorbox}%
\vspace{-0.5em}%
\captionof{figure}{Prompt templates used by the Self-Refine baseline
\citep{madaan2023selfrefine}. The three stages are applied sequentially
per iteration: an initial generation, a self-feedback pass that emits
\texttt{[STOP]} when the solution is deemed correct, and a revision
conditioned on that feedback. Up to four refinement cycles are run on
GSM8K and three on MBPP.}
\label{fig:selfrefine_prompt}
\end{minipage}

\vspace{1em}
\noindent
\begin{minipage}{\textwidth}
\begin{tcolorbox}[colback=gray!5, colframe=gray!60,
    title=\textbf{Self-Reflection Prompt Templates},
    sharp corners=south, width=\textwidth]
    \begin{Verbatim}[fontsize=\footnotesize, commandchars=\\\{\}]
\textbf{-- Stage 1: Initial Answer --}
\textbf{System:} You are a careful math solver. Show all steps clearly.
\textbf{User:} Solve the following math problem step by step.

Problem: \{problem\}

Solution:

\textbf{-- Stage 2: Reflect --}
\textbf{System:} You are a self-critical math solver.
         Diagnose your errors honestly and thoroughly.
\textbf{User:} Your previous answer to the math problem below was incorrect.

Problem: \{problem\}

Incorrect Solution:
\{wrong_solution\}

Provide a structured self-reflection with all five parts:

1. EXPLANATION: Why was the previous solution wrong? Identify the exact
   reasoning or calculation error.
2. ERROR KEYWORDS: List 3-5 concise keywords describing the error type
   (e.g., 'arithmetic error', 'wrong formula', 'misread problem').
3. CORRECT SOLUTION: Solve the problem correctly step by step.
4. INSTRUCTIONS: Write step-by-step instructions for solving this type
   of problem correctly in future.
5. GENERAL ADVICE: What general principle should be remembered to avoid
   this kind of mistake?

Self-Reflection:

\textbf{-- Stage 3: Re-Answer --}
\textbf{System:} You are a careful math solver. Apply your reflection to get it right.
\textbf{User:} Using your self-reflection as a guide, solve the math problem correctly.

Problem: \{problem\}

Your Self-Reflection:
\{reflection\}

Now provide the correct solution, following your own instructions above:

Final Solution:
    \end{Verbatim}
\end{tcolorbox}%
\vspace{-0.5em}%
\captionof{figure}{Prompt templates used by the Self-Reflection baseline
\citep{10852426}. The initial answer is evaluated against
the verifier; if incorrect, a structured five-part reflection is generated
and injected into the re-answer prompt. This is a single-pass operation
with no further iteration.}
\label{fig:selfreflection_prompt}
\end{minipage}


\newpage
\section*{NeurIPS Paper Checklist}

\begin{enumerate}

\item {\bf Claims}
    \item[] Question: Do the main claims made in the abstract and introduction accurately reflect the paper's contributions and scope?
    \item[] Answer: \answerYes{} 
    \item[] Justification:  The abstract and introduction claim four contributions: (1) a principled interventional framework for step-level causal attribution, (2) a minimal counterfactual repair mechanism producing validated contrastive supervision pairs, (3) empirical evaluation across four diverse benchmarks (GSM8K ~\cite{cobbe2021gsm8k}, MBPP~\cite{austin2021program}, SealQA Hard~\cite{pham2025sealqa}, MedBrowseComp~\cite{chen2025medbrowsecomp}) totaling over 3,000 problems, and (4) a formulation enabling offline preference learning from logged trajectories. All four are directly substantiated in the paper. The abstract's quantitative claims (42.7\% aggregate repair rate, improvements across all benchmarks) are reported in Tables ~\ref{tab:repair_baseline} and  ~\ref{tab:accuracy_baseline}. The claim that causal attribution outperforms heuristic refinement in complex retrieval settings is supported by the SealQA Hard and MedBrowseComp results, where Self-Refine and Self-Reflection produce negative or near-zero deltas while CausalFlow achieves +12.6pp and +30.8pp respectively. Scope limitations noted in the abstract, such as the dependency on re-execution feasibility and LLM proposal quality, are discussed in the Limitations section.

\item {\bf Limitations}
    \item[] Question: Does the paper discuss the limitations of the work performed by the authors?
    \item[] Answer: \answerYes{} 
    \item[] Justification: The paper includes a dedicated Limitations section that discusses five specific limitations: (1) intervention quality depends on the LLM's ability to generate meaningful corrective proposals, meaning causally responsible steps may be under-identified if proposals are weak; (2) CRS computation requires re-execution of affected trace segments, introducing computational overhead that may require approximation strategies for very long or tool-intensive traces; (3) retrieval-driven failures cannot be corrected through local reasoning modification when essential information is unavailable; (4) CausalFlow's structured trace-logging configuration reduces initial accuracy relative to plain chain-of-thought baselines on GSM8K (75.0\% vs 88.1\% for Direct), representing a deployment cost not present in simpler baselines; and (5) causal attribution quality depends on accurate dependency annotations, where incomplete or noisy sequential chains may weaken intervention semantics.

\item {\bf Theory assumptions and proofs}
    \item[] Question: For each theoretical result, does the paper provide the full set of assumptions and a complete (and correct) proof?
    \item[] Answer: \answerNA{} 
    \item[] Justification:  The paper does not include formal theorems or proofs. 
The technical content consists of Definition~\ref{def:crs} (Causal 
Responsibility Score), operational formulas (Equations~\ref{eq:crs}, 
\ref{eq:minimality}, \ref{eq:repair}, and \ref{eq:consensus} for CRS, 
Minimality, Repair Selection, and Consensus), and Algorithm~\ref{alg:crs} 
(Causal Responsibility Scoring), all of which are procedural rather than 
theoretical in nature. The framework is grounded in established 
interventionist accounts of actual causation~\cite{halpern2005causes}, 
but CausalFlow itself makes no novel theoretical claims requiring formal 
proof. All results in the paper are empirical.

    \item {\bf Experimental result reproducibility}
    \item[] Question: Does the paper fully disclose all the information needed to reproduce the main experimental results of the paper to the extent that it affects the main claims and/or conclusions of the paper (regardless of whether the code and data are provided or not)?
    \item[] Answer: \answerYes{} 
    \item[] Justification: Full experimental details are provided in Appendix~\ref{app:Appendix_A}, 
including agent configurations for all four 
benchmarks (~\cite{cobbe2021gsm8k, austin2021program, pham2025sealqa, 
chen2025medbrowsecomp}), model identifiers, temperatures, intervention 
protocol (K=3, temperature-based sampling), dataset sources with 
HuggingFace config strings, validation methods, and prompt templates 
for CausalFlow (Figures~\ref{fig:intervention_prompt}, ~\ref{fig:causal_prompt_templates}) and baselines~\cite{madaan2023selfrefine, 
10852426} (Figures ~\ref{fig:selfrefine_prompt},~\ref{fig:selfreflection_prompt}). The anonymized codebase is 
publicly available at \url{https://anonymous.4open.science/r/CausalFlow-6B2C}.

\item {\bf Open access to data and code}
    \item[] Question: Does the paper provide open access to the data and code, with sufficient instructions to faithfully reproduce the main experimental results, as described in supplemental material?
    \item[] Answer: \answerYes{} 
    \item[] Justification: The complete implementation, including all agent configurations, causal attribution algorithms, repair generation modules, and evaluation scripts, is released at \url{https://anonymous.4open.science/r/CausalFlow-6B2C} (Appendix~\ref{app:code_and_reproduce}). All datasets used (GSM8K, MBPP, SealQA Hard, MedBrowseComp) are publicly available on HuggingFace.

\item {\bf Experimental setting/details}
    \item[] Question: Does the paper specify all the training and test details (e.g., data splits, hyperparameters, how they were chosen, type of optimizer) necessary to understand the results?
    \item[] Answer: \answerYes{} 
    \item[] Justification: Section~\ref{sec:agent_config} specifies all agent configurations, model checkpoints, and temperature settings for each benchmark. Appendix~\ref{app:GSM8K_setup} -~\ref{app:medbrowse_setup} provides full details, including dataset splits, HuggingFace config strings, tool configurations, prompt templates, and validation methods. Baseline configurations are described in Section~\ref{sec:agent_config} and Appendix~\ref{sec:baseline_prompts}.

\item {\bf Experiment statistical significance}
    \item[] Question: Does the paper report error bars suitably and correctly defined or other appropriate information about the statistical significance of the experiments?
    \item[] Answer: \answerNo{} 
    \item[] Justification: Due to the high computational cost of running full experiments 
across 3,004 problems with K=3 interventions per step, full error bars 
across multiple runs are not reported for the main results (Tables ~\ref{tab:repair_baseline} and ~\ref{tab:accuracy_baseline}). However, a stochasticity ablation on a 50-problem 
GSM8K~\cite{cobbe2021gsm8k} subset (Table ~\ref{tab:stoch-subset}, Table ~\ref{tab:stoch-runs}, and Section ~\ref{sec:ablations}) reports 
repair rate std.\ dev.\ of 6.81pp and post-repair accuracy std.\ dev.\ 
of 1.63pp across three independent runs, indicating stable outcomes under 
repeated sampling. Wilson's 95\% confidence intervals are reported for LLM 
judge precision in Table~\ref{tab:judge} for SealQA Hard~\cite{pham2025sealqa} and 
MedBrowseComp~\cite{chen2025medbrowsecomp}.

\item {\bf Experiments compute resources}
    \item[] Question: For each experiment, does the paper provide sufficient information on the computer resources (type of compute workers, memory, time of execution) needed to reproduce the experiments?
    \item[] Answer: \answerYes{} 
    \item[] Justification: Appendix ~\ref{app:runtime_analysis} reports wall-clock runtime and approximate API cost for each benchmark: GSM8K (12 hours, \$31.56), MBPP (6 hours, 
\$9.72), SealQA Hard (1 hour, \$16.36), and MedBrowseComp (2 hours, 
\$27.05). Ablation experiments (Section~\ref{sec:ablations}) incurred an additional 
\$45.31, bringing the total API cost to approximately \$130.00. All LLM 
calls were made via the OpenRouter API~\cite{openrouter}. The average 
API cost per successful repair ranged between approximately \$0.05 and 
\$0.15 across datasets.
    
\item {\bf Code of ethics}
    \item[] Question: Does the research conducted in the paper conform, in every respect, with the NeurIPS Code of Ethics \url{https://neurips.cc/public/EthicsGuidelines}?
    \item[] Answer: \answerYes{} 
    \item[] Justification: The research conforms with the NeurIPS Code of Ethics. The paper studies failure diagnosis and repair in LLM agents using publicly available benchmarks, involves no human subjects, and the Impact Statement (Section~\ref{sec:Impact}) discusses both positive applications and potential misuse risks.

\item {\bf Broader impacts}
    \item[] Question: Does the paper discuss both potential positive societal impacts and negative societal impacts of the work performed?
    \item[] Answer: \answerYes{} 
    \item[] Justification: The Impact Statement (Section~\ref{sec:Impact}) discusses positive societal impacts, including safer deployment of LLM agents in education, programming assistance, information retrieval, and healthcare support, as well as the potential negative impact of systematic failure mode analysis being misused to probe system weaknesses or optimize adversarial inputs. The authors note this risk does not introduce new attack capabilities beyond standard evaluation practices.
    
\item {\bf Safeguards}
    \item[] Question: Does the paper describe safeguards that have been put in place for responsible release of data or models that have a high risk for misuse (e.g., pre-trained language models, image generators, or scraped datasets)?
    \item[] Answer: \answerNA{} 
    \item[] Justification: The paper releases a debugging and repair framework and evaluation code, not pre-trained models, image generators, or scraped datasets. The anonymized codebase poses no elevated misuse risk beyond standard ML research tools.

\item {\bf Licenses for existing assets}
    \item[] Question: Are the creators or original owners of assets (e.g., code, data, models), used in the paper, properly credited and are the license and terms of use explicitly mentioned and properly respected?
    \item[] Answer: \answerYes{} 
    \item[] Justification: All datasets are cited with their original papers: GSM8K~\cite{cobbe2021gsm8k}, MBPP~\cite{austin2021program}, SealQA Hard~\cite{pham2025sealqa}, and MedBrowseComp~\cite{chen2025medbrowsecomp}, and are loaded from their public HuggingFace repositories with config strings specified in Appendix~\ref{app:GSM8K_setup}-~\ref{app:medbrowse_setup}. Baseline methods (Self-Refine ~\cite{madaan2023selfrefine}, Self-Reflection ~\cite{10852426}) are similarly cited. The Serper API~\cite{serper2025} and OpenRouter API~\cite{openrouter} are referenced in Appendix ~\ref{app:runtime_analysis} and Appendix ~\ref{app:sealQA_setup} respectively.

\item {\bf New assets}
    \item[] Question: Are new assets introduced in the paper well documented and is the documentation provided alongside the assets?
    \item[] Answer: \answerYes{} 
    \item[] Justification: The CausalFlow codebase is released as a new asset at \url{https://anonymous.4open.science/r/CausalFlow-6B2C} (Appendix~\ref{app:code_and_reproduce}), including all agent configurations, attribution algorithms, repair modules, and evaluation scripts with documentation and example traces.

\item {\bf Crowdsourcing and research with human subjects}
    \item[] Question: For crowdsourcing experiments and research with human subjects, does the paper include the full text of instructions given to participants and screenshots, if applicable, as well as details about compensation (if any)? 
    \item[] Answer: \answerNA{} 
    \item[] Justification: The paper does not involve crowdsourcing or research with human subjects. The manual audit in Table~\ref{tab:judge} was performed by the authors as a quality check on LLM-graded outputs, not as a crowdsourcing study.

\item {\bf Institutional review board (IRB) approvals or equivalent for research with human subjects}
    \item[] Question: Does the paper describe potential risks incurred by study participants, whether such risks were disclosed to the subjects, and whether Institutional Review Board (IRB) approvals (or an equivalent approval/review based on the requirements of your country or institution) were obtained?
    \item[] Answer: \answerNA{} 
    \item[] Justification: The paper does not involve crowdsourcing or research with human subjects and therefore does not require IRB approval.

\item {\bf Declaration of LLM usage}
    \item[] Question: Does the paper describe the usage of LLMs if it is an important, original, or non-standard component of the core methods in this research? Note that if the LLM is used only for writing, editing, or formatting purposes and does \emph{not} impact the core methodology, scientific rigor, or originality of the research, declaration is not required.
    \item[] Answer:  \answerYes{} 
    \item[] Justification: LLMs are a core component of the CausalFlow methodology. Section~\ref{sec:agent_config} and Appendix~\ref{app:intervention_prompting} -~\ref{app:medbrowse_setup} fully describe LLM usage across all pipeline components: intervention proposal generation (Gemini 2.0 Flash Lite, GPT-5 Chat, Gemini 3 Flash Preview), predictive re-execution for outcome validation on GSM8K, SealQA Hard, and MedBrowseComp, multi-agent validation (Section~\ref{sec:multi_agent_validation}), and LLM-based grading for browsing benchmarks.

\end{enumerate}

\end{document}